\documentclass[10pt, a4paper]{article}

\usepackage[]{lrec-coling2024} 
\usepackage{times}
\usepackage{latexsym}
\usepackage[T1]{fontenc}
\usepackage{subfig}
\usepackage[utf8]{inputenc}
\usepackage{inconsolata}
\usepackage{amsmath}
\usepackage{microtype}

\usepackage{amssymb}
\usepackage{booktabs}
\usepackage{xspace}
\usepackage{xcolor}
\usepackage{todonotes}
\usepackage{hyperref}
\usepackage{multicol}
\usepackage{multirow}
\usepackage{enumitem}
\usepackage{xcolor,soul,colortbl}

\title{Exploring Geometric Representational Disparities Between \\   Multilingual and Bilingual Translation Models}

\name{Neha Verma$^{1}$, Kenton Murray$^{1,2}$, Kevin Duh$^{1,2}$ } 

\address{ $^{1}$Center for Language and Speech Processing \\  $^{2}$Human Language Technology Center of Excellence \\\{nverma7, kenton\}@jhu.edu, kevinduh@cs.jhu.edu \\ }

\abstract{ 
Multilingual machine translation has proven immensely useful for both parameter efficiency and overall performance across many language pairs via complete multilingual parameter sharing. 
However, some language pairs in multilingual models can see worse performance than in bilingual models, especially in the one-to-many translation setting. 
Motivated by their empirical differences, we examine the geometric differences in representations from bilingual models versus those from one-to-many multilingual models.
Specifically, we compute the isotropy of these representations using intrinsic dimensionality and IsoScore, in order to measure how the representations utilize the dimensions in their underlying vector space.
Using the same evaluation data in both models, we find that for a given language pair, its multilingual model decoder representations are consistently less isotropic and occupy fewer dimensions than comparable bilingual model decoder representations.
Additionally, we show that much of the anisotropy in multilingual decoder representations can be attributed to modeling language-specific information, therefore limiting remaining representational capacity. 
 \\ \newline }

\begin{document}

\maketitleabstract

\section{Introduction}

Recent advances in multilingual machine translation have led to better parameter efficiency and language transfer by simultaneously modeling multiple language pairs \cite{firat-etal-2016-multi, ha-etal-2016-toward}. 
Some work has even proven the viability of performing zero-shot translation between language pairs for which there may be very little to no bitext \cite{johnson-etal-2017-googles, zhang-etal-2020-improving}.
However, multilingual translation systems with complete parameter sharing can suffer from interference, or reduced performance for some language pairs versus a comparable bilingual baseline \cite{aharoni-etal-2019-massively, arivazhagan2019massively}. 

Previous work has hypothesized that limited modeling capacity is a major contributor to reduced performance in multilingual models  \cite{aharoni-etal-2019-massively, zhu-etal-2021-counter-interference, conneau-etal-2020-unsupervised}. 
Some prior work shows this bottleneck phenomenon empirically by evaluating bilingual versus multilingual model performance across different model and data sizes  \cite{zhang-etal-2020-improving, shaham_causes}.
Besides capacity, the direction of translation can also dictate how much interference occurs in multilingual models; one-to-many translation systems suffer more from interference compared to multilingual translation model types \cite{wang-etal-2018-three, arivazhagan2019massively, fernandes2023scaling}. 
Therefore, in this work, we focus on one-to-many multilingual translation systems.

\begin{figure}[!ht]
    \centering
    \includegraphics[width=7cm]{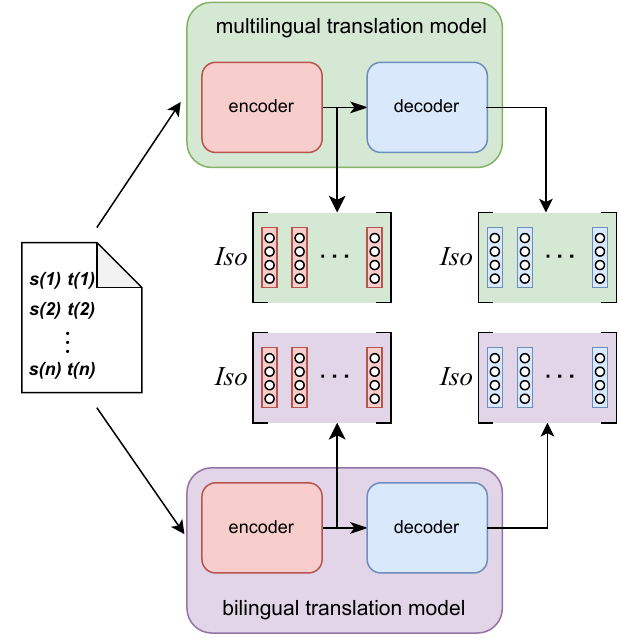}
    \caption{Schematic of our hidden space utilization comparisons. We extract final layer representations from both a  bilingual model and a multilingual model on the same set of parallel sentences. We compute the isotropy of these representations (\textit{Iso}), and compare the two models. }
    \label{fig:page-one}
\end{figure}

Despite trends pointing towards performance differences between bilingual and multilingual translation systems, especially in those with a multilingual decoder, it still unclear \textit{how} these systems may be performing differently.
To this end, we systematically compare the behavior of one-to-many translation models to their bilingual counterparts. 
Specifically, we examine the geometry of model representations from both types of models and compare them directly. 
We ask the following: 
(1) How does the ambient space utilization of model representations differ between bilingual models and one-to-many models? 
(2) If space utilization differs, what might be driving these differences?

We measure space utilization using IsoScore and intrinsic dimensionality (ID), which are two metrics that determine how uniformly a point cloud utilizes the dimensions of its underlying vector space, or its isotropy \cite{fukunaga-olsen, rudman-etal-2022-isoscore}.   

We compute the isotropy of representations on the same set of sentence pairs across model types so that their scores are directly comparable, and summarize our method in Figure \ref{fig:page-one}. 
We observe the following in our comparison:
\begin{itemize}[itemsep=-0.2em ]
    \item Across different data resource levels and different source-target language pairs, the isotropy of one-to-many decoder representations for a given source-target pair is reduced as contrasted with decoder representations in a comparable bilingual model. 
    \item 
    Source-side representation capacity improves slightly in one-to-many models over bilingual models. However, the extent of this encoder capacity improvement is smaller than the extent of the decoder capacity reduction.
    \item With further analysis, we find that reduced space utilization in multilingual decoder representations seems driven by language-specific information occupying much of the available representation space. Single language decoders, however, do not have to distinguish this language-specific information.
\end{itemize}

While most previous work has observed empirical differences between bilingual and multilingual models and some of its potential causes, our work characterizes the differences between bilingual and multilingual models in terms of their internal model representations. Our results could inform alternative approaches on current multilingual modeling design, especially in models that cover multiple target languages. 

\section{Analysis of Model Representations}
\label{sec:analysis}

\subsection{Model Representation Space Utilization}

In this work, we investigate the difference between our model types via the geometry of final and intermediate layer representations. 
Specifically, we are interested in how well these representations utilize the dimensions of the vector space they lie in.  
If a set of representations has very high variance across a few dimensions, and little to no variance spread across the remaining dimensions, this set is said to have low isotropy, or anisotropy. 

Because a one-to-many model has to accommodate multiple languages in its decoder, we hypothesize that our multilingual models have less representational capacity than bilingual models for a given language pair. 
Therefore, we turn to examining the isotropy of representations produced from both a bilingual model and a multilingual model on a set of parallel sentences. 
Since our experiments keep the hidden dimension fixed across all models, and the representations are computed from the same data, these two sets of hidden vectors are directly comparable. 
In this setting, if one set of representations uses more ambient vector space compared to the other set, we can say that the first set is using more of its representational capacity. 

\subsection{Computing Isotropy}
\label{sec:computing_isotropy}
In computing the space utilization of model representations, we first compute the sequence of hidden states across tokens.
For a given source target pair $(\mathbf{x}, \mathbf{y}$), a forward pass through the encoder gives ${h}_{\text{enc}}(\mathbf{x}) = (v_1, v_2, \hdots, v_{|\mathbf{x}|})$, and through the decoder gives ${h}_{\text{dec}}(\mathbf{x},\mathbf{y}) = (w_1, w_2, \hdots, w_{|\mathbf{y}|})$

We compute the isotropy of these model representations at a sentence level. For converting encoder and decoder hidden state sequences into single vectors, we mean pool all non-padding tokens over the token dimension \cite{li-etal-2020-sentence, kudugunta-etal-2019-investigating}.
Isotropy, formally, is a measure of how uniformly the variance of a dataset is spread across its vector dimensions.

The isotropy metrics used in this work are  intrinsic dimensionality (ID) as computed by the PCA Fukunaga-Olsen algorithm \cite{fukunaga-olsen} and  IsoScore \cite{rudman-etal-2022-isoscore}. 
PCA Fukunaga-Olsen is a straightforward method to estimate the ID of a dataset based on a linear PCA decomposition of the data. This method is simple, robust to large samples, and handles high dimensionality, which is important for our hidden vector setting \cite{bac2021scikit}. The PCA-FO ID algorithm computes the following, for threshold $D_e \in [0,1)$ and original dimensionality $n$: 
\begin{enumerate}
\itemsep-0.3em 
    \item  Compute PCA of the dataset $X \subseteq \mathbb{R}^n $: $\text{cov}(X) = V\Lambda V^T$
    \item Compute normalized eigenvalues $\lambda_i = \lambda_i / \lambda_1$
    \item return $\text{count}(\lambda_i > D_e) / n$
\end{enumerate}
In this work, we use $D_e = 0.05$. 

IsoScore is a similar metric that uses the diagonal of the covariance matrix of PCA-transformed points in order to measure how many dimensions are used and how uniformly the dimensions are used. 
Previous works on representation isotropy have used other metrics, like average cosine similarity or partition scores \cite{mu2018all, ethayarajh-2019-contextual}, but \citet{rudman-etal-2022-isoscore} found that these methods do not stand up to thorough validity testing, like mean agnosticism or rotational invariance. 

More formally, IsoScore computes the following:
\begin{enumerate}
    \itemsep-0.3em 
    \item  Reorient dataset $X \subseteq \mathbb{R}^n $ with PCA: $X^{\text{PCA}}$
    \item  Compute the diagonal covariance matrix of $X^{\text{PCA}} \in \mathbb{R}^n$, denoted as $\Sigma_D$. \footnote{PCA guarantees no off-diagonal covariance elements.}
    \item  Normalize the variance diagonal to be: $\hat{\Sigma}_D := \sqrt{n} \frac{\Sigma_D}{\lVert \Sigma_D \rVert}$
    \item Compute the distance between the covariance diagonal and the identity matrix, which reflects ideal isotropy: $\delta(X) := \frac{\lVert \hat{\Sigma}_D - \mathbf{1} \rVert}{\sqrt{2(n- \sqrt{n})}}$
    \item Use $\delta(X)$ to compute the percentage of dimensions isotropically utilized. \\$\phi(X) = (n - \delta(X)^2(n - \sqrt{n}))^2 / n^2$
\end{enumerate}

\begin{figure}[!h]
    \centering
    \includegraphics[width=\columnwidth]{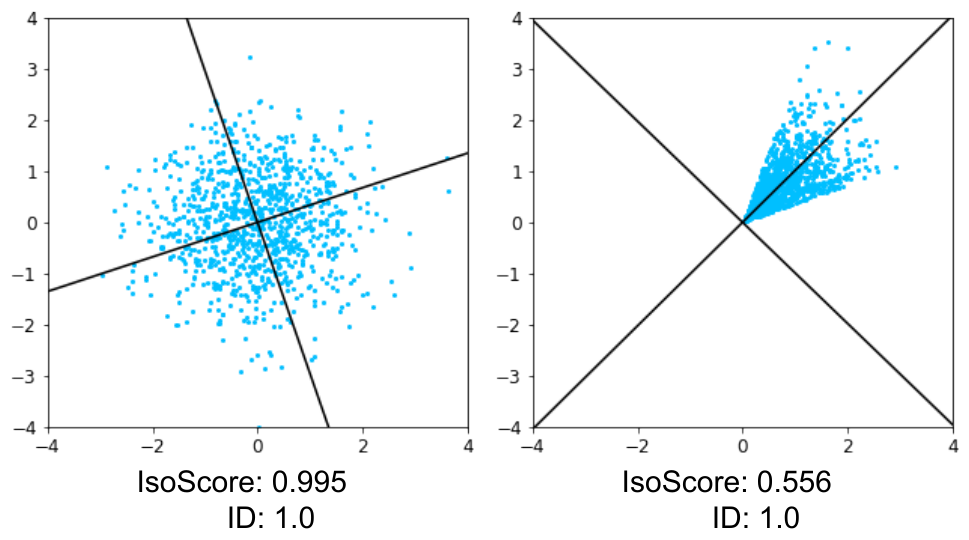}
    \caption{Depictions of 2D point clouds, their principal components, and their computed IsoScores and IDs. The left point cloud has high IsoScore due to even variance spread across principal components, but the right has lower IsoScore due to uneven variance spread. Both clouds have an ID of 1.0 as ID is less sensitive to variance spread.}
    \label{fig:page-two}
\end{figure}

The final range of $\phi(X)$ is linearly rescaled to span the interval $[0,1]$, resulting in the IsoScore.
More details and motivation behind the metric can be found in the original paper \cite{rudman-etal-2022-isoscore}. We detail an example of point clouds and their respective IsoScores and IDs in Figure \ref{fig:page-two}. 

The main difference between IsoScore and ID is that IsoScore accounts for evenness of variance spread among the dimensions, whereas ID only computes a variance threshold. In our Figure \ref{fig:page-two} example, the ID of these point clouds is both 1.0, meaning that all dimensions are utilized, but the IsoScore captures more fine-grained detail about \textit{how} the dimensions are being used.

In our work, we compute IsoScores and the ID of several sets of model representations for comparison. 
We begin with a multilingual model that translates language pairs $s \rightarrow \{t_1, t_2, ..., t_n\}$, a bilingual model that translates only  $s \rightarrow t_k$, and a set of sentences $\{s(i), t_k(i)\}$. 
For both models, we compute the isotropy using one of our metrics, of $X_{\text{enc}} = \{h_{\text{enc}}(s(i)) :\forall i\}$ and $X_{\text{dec}} = \{h_{\text{dec}}(s(i), t_k(i)) :\forall i\}$. These values are labelled $Iso(X_{\text{enc}}^{\text{multi}}(s, t_k)), Iso(X_{\text{dec}}^{\text{multi}}(s, t_k))$ and $Iso(X_{\text{enc}}^{\text{bi}}(s, t_k)), Iso(X_{\text{dec}}^{\text{bi}}(s, t_k))$. Additionally, to observe the overall behavior of our multilingual models, we compute the isotropy of hidden states from all covered language pairs, resulting in $Iso(X_{\text{enc}}^{\text{multi}}(s, \bigcup\limits_{j} t_j)), Iso(X_{\text{dec}}^{\text{multi}}(s, \bigcup\limits_{j}  t_j))$. 

\section{Experimental Setup}

\subsection{Trilingual Models}

In order to control for the effects of language similarity, we experiment with trilingual models that translate from English to two languages, keeping one of the target languages fixed \cite{xincurrent, fernandes2023scaling, shaham_causes}. 
Specifically, we look at trilingual models with English as a source language, and 2 target languages. 
We use Russian (ru) as a fixed target, and vary the 3 other target languages: Chinese (zh), German (de), and Ukrainian (uk). 
These three additional languages have differing degrees of language similarity with Russian; Ukrainian and Russian share a close language family and script, German and Russian share a distant language family and do not share a script, and Russian and Chinese do not share a language family or script.
In summary, we experiment with en-\{ru,zh\}, en-\{ru,de\}, and en-\{ru,uk\} models.

\begin{table*}[ht!]
\begin{center}
\begin{tabular}{llccccccc}
\toprule
& &\multicolumn{2}{c}{WMT-large} & \multicolumn{2}{c}{WMT-small} &  \multicolumn{2}{c}{Multiparallel TED}\\
\cmidrule(lr){3-4} \cmidrule(lr){5-6}   \cmidrule(lr){7-8}  
dataset & lang & train & dev  & train & dev & train & dev\\
\midrule  
\multirow{2}{*}{en-\{ru,de\}}  & en-ru & 98.2M & 2993 & 149k & 2993 & 149k & 1958\\
                            & en-de & 98.2M &  2203 & 149k & 2203 & 149k & 1958\\
\multirow{2}{*}{en-\{ru,uk\}}  & en-ru & 31.5M & 2993 & 67k & 2993 & 67k & 1958\\
                            & en-uk & 31.5M &  997 & 67k & 997 & 67k & 1958\\
\multirow{2}{*}{en-\{ru,zh\}}  & en-ru & 41.1M & 2993 & 161k & 2993 & 161k & 1958\\
                            & en-zh & 41.1M &  3418  & 161k & 3418 & 161k & 1958\\
\bottomrule
\end{tabular}
\end{center}
\caption{Total sentences in each bitext used in our work. We train trilingual models that translate from English into two other languages. We force the WMT-small training split to be the same size as Multiparallel TED for comparability. }
\label{table:data}
\end{table*}

\subsection{Datasets}

Our main experiments use data from previous WMT competitions on general translation.
We use training and development data from the 2022 WMT General Machine Translation task, and describe our WMT data preparation pipeline in Appendix \ref{sec:wmt_cleaning_appendix}.
For validation on our en-\{ru,uk\} multilingual models, we subsample from the WMT22 Russian development set in order to match the size of the Ukrainian set for evenness. However, we perform our analysis on the whole development set.

We additionally use bitext from the Multitarget TED talks, which allow us to investigate the role of multiparallel data in MT representations \cite{duh18multitarget}. 
We filter the Multitarget TED talk training sets to be strictly multiparallel, like their dev and test sets, and henceforth refer to the dataset as multiparallel TED talks. 
To measure the effect of data availability as well as multiparallelism, we subsample our WMT data to match the size of the Multiparallel TED talks. This way, our small WMT set and TED talks can help us study multiparallelism, and our small WMT set and large WMT set can help show the effect of scale on representational capacity. 
Statistics on our datasets are in Table \ref{table:data}.

\subsection{Training Details}
For our bilingual and multilingual translation models, we use the Transformer architecture as implemented by $\texttt{fairseq}$ \cite{transformer, ott-etal-2019-fairseq}. 
For TED and WMT-small experiments, we use the $\texttt{transformer\_iwslt\_de\_en}$ configuration, and for WMT-large experiments, we use a transformer base configuration. 
We use weight tying between decoder input and output embeddings \cite{press-wolf-2017-using, inan2016tying}. For multilingual models, we incorporate target language id tokens prepended to the source sentence \cite{wicks-duh-2022-effects}. 
For all bilingual experiments, we use a joint source-target SentencePiece vocabulary of 16K tokens \cite{sennrich-etal-2016-neural, kudo-richardson-2018-sentencepiece}. 
For all multilingual experiments, we use a joint source-target vocabulary of 32K tokens. These vocabularies have high token overlap, where each multilingual vocabulary contains at least 93\% of the bilingual vocabulary across all languages and datasets. This overlap leads to very similar tokenizations of the sentences in our comparisons. 

For TED and WMT-small experiments, we select the best model checkpoint using validation on BLEU after training for up to 80 epochs.  
For WMT large experiments, we use average validation loss for selection after training up to 240k updates with a batch size of 32k tokens. All outputs are computed using a beam size of 5. 
We report BLEU scores on our dev sets computed with \texttt{sacrebleu} \cite{papineni-etal-2002-bleu, post-2018-call}.

\section{Results}

\begin{table*}[ht]
\begin{center}
\resizebox{\textwidth}{!}{
\begin{tabular}{lcccccccccccc}
\toprule
& & &\multicolumn{5}{c}{WMT-large} & \multicolumn{5}{c}{WMT-small} \\
\cmidrule(lr){4-8} \cmidrule(lr){9-13}  
dataset & langs & type & BLEU & iso-enc & ID-enc & iso-dec & ID-dec & BLEU & iso-enc  & ID-enc & iso-dec & ID-dec\\
\midrule\midrule 
\multirow{5}{*}{en-\{ru,zh\}} & \multirow{2}{*}{en-ru} & multi &  23.7 & \textbf{0.082} & \textbf{0.070} & 0.164 & 0.145 & 20.0 &\textbf{0.104} & \textbf{0.088} & 0.208 & 0.250 \\
&& bi & 23.7 & 0.075 & 0.057 & \textbf{0.192} & \textbf{0.164} & 19.1 & 0.074 & 0.057 & \textbf{0.236} & \textbf{0.285} \\
\cmidrule{2-13}
& \multirow{2}{*}{en-zh}  & multi & 34.5 &\textbf{0.051} & \textbf{0.045} & 0.106 & 0.092 & 28.0 & \textbf{0.070} & \textbf{0.057} & 0.136 & 0.148 \\
&& bi & 36.0  & 0.023 & 0.020 & \textbf{0.142}& \textbf{0.199} & 27.7 & 0.032 & 0.023 & \textbf{0.185} & \textbf{0.201} \\
\cmidrule{2-13}
&  both & multi & - & 0.066 & 0.057 &  0.065 & 0.043 & - & 0.085 & 0.070 & 0.076  & 0.047 \\
\midrule\midrule 
\multirow{4}{*}{en-\{ru,de\}} & \multirow{2}{*}{en-ru} & multi  & 23.8 & \textbf{0.081} & \textbf{0.068} & 0.161 & 0.145 & 19.1 & \textbf{0.112} & \textbf{0.105} & 0.230 & 0.293 \\
&& bi & 23.8 & 0.074 & 0.064 &\textbf{0.189} & \textbf{0.164} & 18.9 & 0.109 &  0.104 & \textbf{0.242} & \textbf{ 0.295}\\
\cmidrule{2-13}
& \multirow{2}{*}{en-de} & multi & 26.4 & \textbf{0.046} & \textbf{0.039} & 0.141 & 0.164 & 18.0 & \textbf{0.076} & \textbf{0.063} & 0.171 & 0.227 \\
&& bi & 28.0 & 0.037 & 0.029 & \textbf{0.191}& \textbf{0.236} & 15.3 & 0.056 & 0.037 & \textbf{0.233} & \textbf{0.311}   \\
\cmidrule{2-13}
&  both & multi & - & 0.049 &0.037 & 0.037	 & 0.021& - &0.070 & 0.049 & 0.070 & 0.039\\
\midrule\midrule 
\multirow{4}{*}{en-\{ru,uk\}} & \multirow{2}{*}{en-ru} & multi & 23.7 &  \textbf{0.053} & \textbf{0.053} & 0.161 & 0.168 & 17.3 & 0.118 & 0.115 & 0.242 & \textbf{0.307} \\
&& bi & 23.8 & 0.031 & 0.029 & \textbf{0.184} & \textbf{0.182} & 15.7 & \textbf{0.123} & \textbf{0.129} & \textbf{0.246} & 0.305 \\
\cmidrule{2-13}
& \multirow{2}{*}{en-uk}  & multi & 26.6 & \textbf{0.086} & \textbf{0.080} & 0.139 & 0.160 & 16.2  & 0.148 & 0.199 & 0.191 & 0.238  \\
&& bi & 27.8 & 0.074 & 0.072 & \textbf{0.195} & \textbf{0.238} & 12.7 & \textbf{0.160} & \textbf{0.213} & \textbf{0.221} & \textbf{0.281}\\
\cmidrule{2-13}
& both  & multi & - & 0.086	& 0.086	& 0.078& 0.051 & - & 0.127&	0.145&0.172 &	0.162\\
\bottomrule\bottomrule
\end{tabular}
}
\end{center}
\caption{Main isotropy results for models trained on WMT data. We report BLEU scores of each model on the appropriate validation set, and IsoScores and intrinsic dimensionalities (ID) for both encoder and decoder sentence representations. We report scores for both language pairs, and in both types of models, bilingual (bi) and multilingual (multi). We bold the higher IsoScore/ID value between each multilingual/bilingual comparison. We additionally report the IsoScore of multilingual model spaces on the entire development set, not separating by language pair (both). }
\label{table:main}
\end{table*}

 \begin{figure*}[h!]%
    \centering
    \subfloat[\centering WMT-large, en-ru]{{\includegraphics[width=8cm]{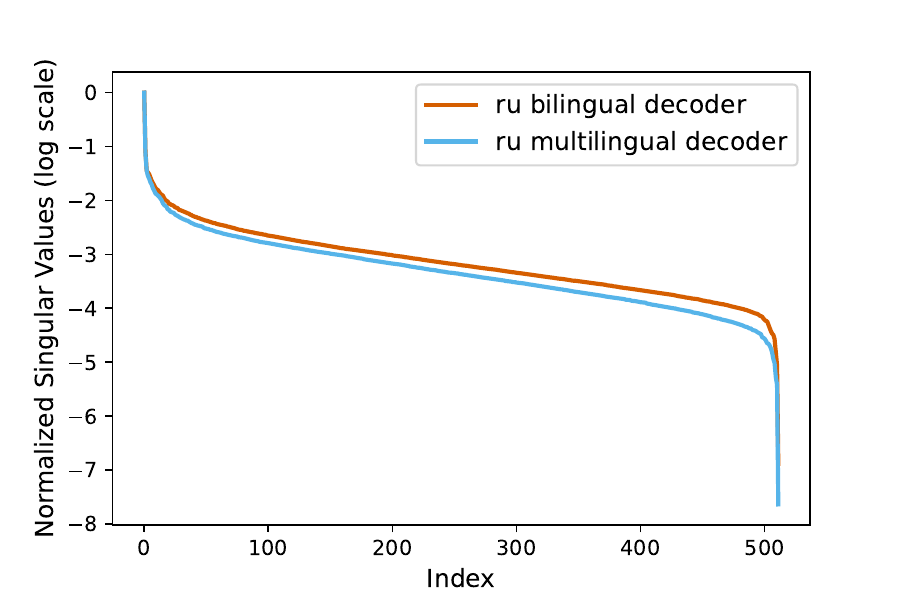} }}%
    \subfloat[\centering WMT-large, en-zh]
    {{\includegraphics[width=8cm]{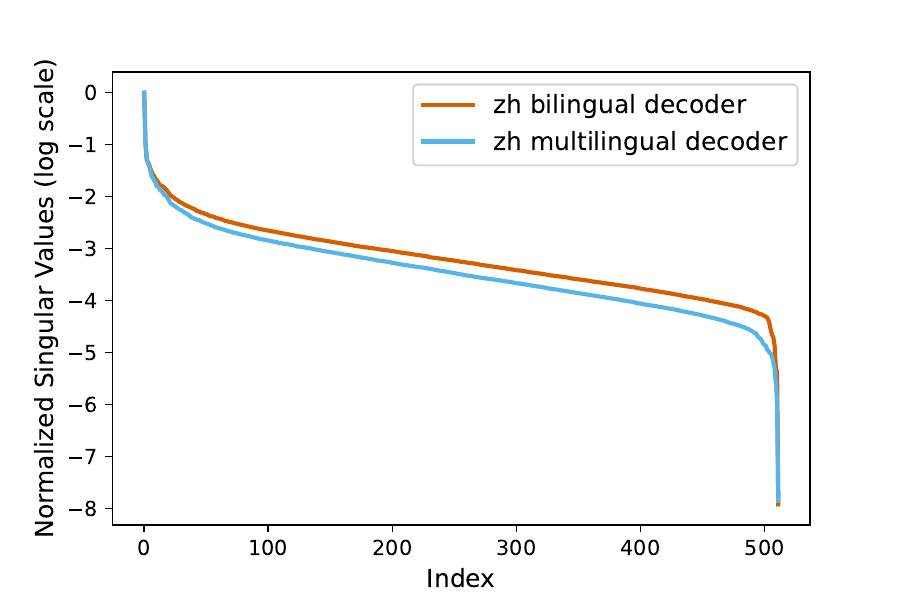} }} \\
    \caption{Semi-log plots of normalized singular values from SVD of bilingual decoder hidden states and multilingual decoder hidden states for the WMT-large en-\{ru,zh\} model. The spectra of bilingual decoder hidden states are better balanced than those of multilingual decoder hidden states. We use a semi-log scale for visibility.}%
    \label{fig:eigenvalues}%
\end{figure*}

\subsection{Multilingual decoder capacity reduction}
We find that across our language pair settings and across our dataset sizes, representations from bilingual model decoders are more isotropic than multilingual model decoder representations.
In Table \ref{table:main}, we see that for all trilingual settings, and for both WMT-small and WMT-large, bilingual decoder isotropy scores are larger than those of multilingual models for the same language pair. For example, in the WMT-large en-\{ru,zh\} dataset, the IsoScore of multilingual decoder representations (iso-dec) for Russian is 0.164 and Chinese is 0.106, but in their respective bilingual models, these values jump to 0.192 for Russian and 0.142 for Chinese. 

Additionally, we plot the singular values from  the singular value decomposition (SVD) of the hidden states of one of our multilingual model decoders and its corresponding two bilingual model decoders in Figure \ref{fig:eigenvalues}.
We see that the spectra of the bilingual model decoder hidden states are more balanced than those of from the multilingual model, as they do not drop off in value as quickly as the multilingual singular values. 
This additionally demonstrates that the bilingual decoder hidden states have better distribution of variance across its dimensions.

Because these representations are computed from the same set of source-target sentences, and only the model types differ, the multilinguality of the one-to-many decoder must be contributing its reduced representational capacity for the  source-target pair. 
In this case, modeling language-specific information in each decoder pass may be occupying much of the multilingual decoder state space.  
We explore this hypothesis further in Section \ref{sec:dec_lang_sep}. 

\subsection{Multilingual encoder capacity increase}

\begin{figure}[h!]%
    \centering
    \subfloat[\centering WMT-large]{{\includegraphics[width=6.7cm]{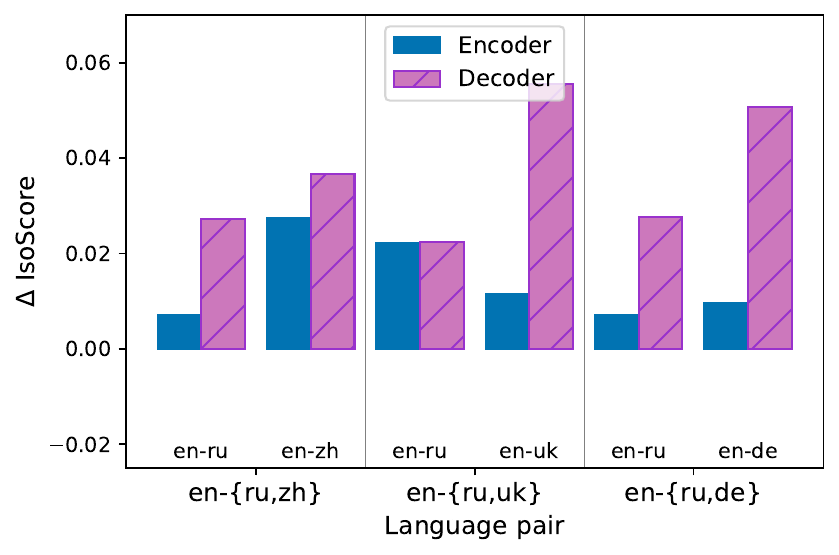} }}%
    \qquad
    \subfloat[\centering WMT-small]
    {{\includegraphics[width=6.7cm]{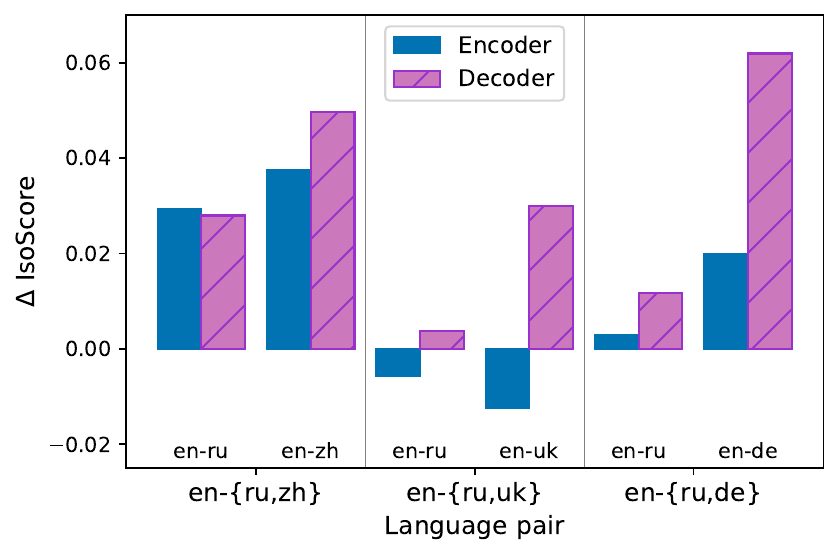} }} \\
    \caption{$\Delta$IsoScore values comparing the extent of the observed encoder isotropy increase ($Iso(X_{\text{enc}}^{\text{multi}}) - Iso(X_{\text{enc}}^{\text{bi}}) $) to the extent of the observed isotropy decrease ($Iso(X_{\text{dec}}^{\text{bi}} )- Iso(X_{\text{dec}}^{\text{multi}}) $) in our multilingual models, compared to their bilingual counterparts. Overall, the extent of the decoder isotropy decrease is larger than that of the encoder increase. }%
    \label{fig:enc_v_dec}%
\end{figure}

In encoder representation spaces, we see an opposite effect, although less pronounced. 
In both en-\{ru,zh\} and en-\{ru,de\} models, across small and large data availability, multilingual encoders tend to have greater isotropy among representations than bilingual model encoders. 
However, the one exception is the WMT-small en-\{ru-uk\} model. 
Results comparing this increase in encoder capacity to the decrease in decoder capacity in multilingual models, compared to their bilingual counterparts, are summarized in Figure \ref{fig:enc_v_dec}. 

Comparing multilingual encoder isotropy separated by language versus the isotropy of the whole multilingual encoder space (Table \ref{table:main}), we see that the difference in scores is not very large. 
This could indicate that the multilingual encoder space is benefiting from sharing across the English sources from both language pairs in our multilingual dataset.

\subsection{Effects of training scale}

In comparing IsoScore results on WMT-small vs WMT-large setups, we see that in a larger scale, there is consistently less space utilization in both multilingual and bilingual models.
This occurs consistently in the decoder space, and in almost all settings in the encoder space. 
Both models have the same hidden dimension $d=512$, and differ only in their feed-forward dimension and attention heads.
Even among the overall multilingual isotropy scores (setting labeled `both' in Table \ref{table:main}), WMT-large representations have smaller isotropy values than WMT-small representations in almost all language settings.

The observed increase in anisotropy with larger training scale is closely related to the representation degeneration problem reported in previous literature \cite{gaorepresentation}. This phenomenon describes a tendency towards anisotropy of the final softmax layer $W$ in natural language generation models, due to a frequency bias affecting output token embedding updates. 
With more training updates, this frequency bias causes output token embeddings to become more anisotropic.
In our case, we see a similar degeneration with final hidden states, which are closely related to the softmax layer given the output distribution computation $y = \text{softmax}(h^TW)$ where $h$ is our final hidden vector. 

In terms of performance, we note that only the WMT-large BLEU scores see a reduction or no improvement in the multilingual case; it is known that measurable interference does not generally occur much at a smaller data scale \cite{shaham_causes}. 

\subsection{Multi-way parallelism}

We report results on the Multiparallel TED Talks in Table \ref{table:ted}. 
In this setting, we find that our results on increased isotropy of multilingual source-side representations still holds in a majority of cases, even though the source-side sentences are identical across our two language pairs in the trilingual model. 
This is a strong indication that in one-to-many models, source-side representations benefit from a shared source embedding space, and do not separate much based on target language. 

On the other hand, our results on decreased decoder capacity do not hold in all language settings in our multiparallel model.
An isotropy increase occurs over bilingual models to a small extent for our en-\{ru,de\} model, and a larger one for our en-\{ru,uk\} model, where the target languages share a script. 
However, the isotropy of our entire decoder multilingual space is still relatively low. 
This indicates that although there is still separation in the decoder space by language, each language's representation cluster in the decoder space is still more locally isotropic than its bilingual counterpart. 

We test our TED model on our WMT test sets for direct comparability to our other models. 
Full results can be found in Appendix \ref{sec:appendix_ted_wmt}.
We see that results are mostly consistent for multilingual encoder isotropy improvement. 
For multilingual decoder isotropy, we see similar results with respect to language relatedness --- 
bilingual decoder representations are more anisotropic than their multilingual counterparts for en-\{ru,zh\}, similar for en-\{ru,de\}, and the opposite for en-\{ru,uk\}, where the target languages are most related.

\begin{table}[h!]
\begin{center}
\resizebox{0.48\textwidth}{!}{
\begin{tabular}{lcccccc}
\toprule
& &\multicolumn{5}{c}{Multiparallel TED} \\
\cmidrule(lr){3-7}
 langs & type & BLEU & iso-enc & ID-enc & iso-dec & ID-dec \\
\midrule\midrule
\multirow{2}{*}{en-ru} & multi & 16.0 & \textbf{0.135} & \textbf{0.133} &  0.253 & 0.313  \\
& bi & 15.5 & 0.130 & 0.119 &\textbf{ 0.284} & \textbf{0.348} \\
\cmidrule{2-7}
\multirow{2}{*}{en-zh}  & multi & 19.3 & 0.122 & 0.113 & 0.244 & 0.305  \\
& bi & 18.8 & \textbf{0.125} & \textbf{0.125} & \textbf{0.277} & \textbf{0.338} \\
\cmidrule{2-7}
 & both & - & 0.138 & 0.137 & 0.104 & 0.063 \\
\midrule\midrule
 \multirow{2}{*}{en-ru} & multi & 15.8 &\textbf{ 0.108} & \textbf{0.094} & \textbf{0.261} & \textbf{0.326} \\
& bi &  15.3 & 0.097 & 0.098 & 0.250 & 0.309 \\
\cmidrule{2-7}
\multirow{2}{*}{en-de}  & multi & 26.1 & \textbf{0.104} & \textbf{0.088}  & \textbf{0.258} & \textbf{0.320}  \\
& bi & 25.2 & 0.073 & 0.066 & 0.247 & 0.287 \\
\cmidrule{2-7}
& both & -& 0.108 & 0.094 & 0.116 & 0.072  \\
\midrule\midrule
\multirow{2}{*}{en-ru} & multi  &  13.5 & \textbf{0.127} & 0.139 & \textbf{0.248} &  \textbf{0.305}\\
& bi & 12.2 & 0.124 & \textbf{0.145} & 0.222 & 0.260 \\
\cmidrule{2-7}
\multirow{2}{*}{en-uk} & multi &  16.8 & \textbf{0.128} & 0.141 & \textbf{0.244} & \textbf{0.299} \\
& bi &  15.6 & 0.124 & \textbf{0.152} & 0.201 & 0.238 \\
\cmidrule{2-7}
 & both & - &0.130  & 0.143 & 0.173 & 0.168 \\
\bottomrule\bottomrule
\end{tabular}
}
\end{center}
\caption{Isotropy results on the encoder and decoder sentence representations from our Multiparallel TED model, tested on the Multiparallel TED development set.}
\label{table:ted}
\end{table}

\begin{figure}[h!]%
    \centering
    \subfloat[\centering WMT-large]{{\includegraphics[width=7cm]{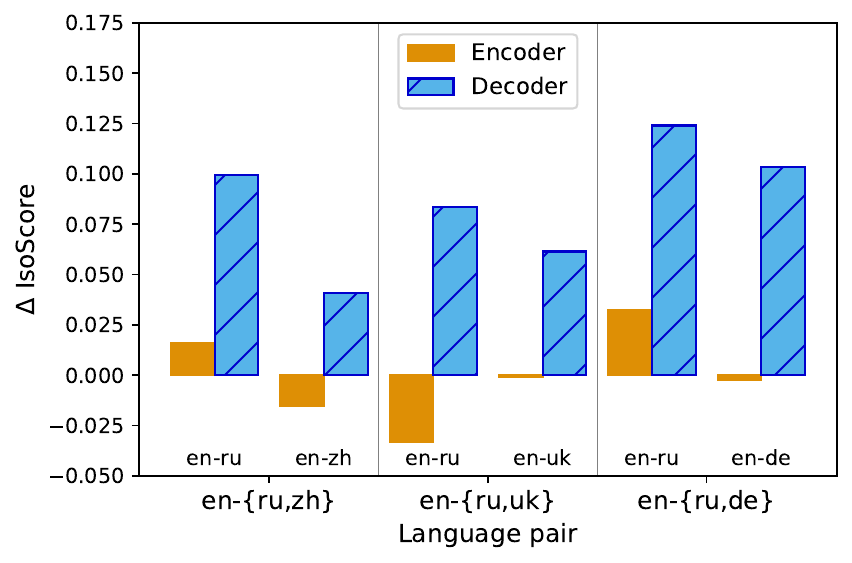} }}%
    \qquad
    \subfloat[\centering WMT-small]{{\includegraphics[width=7cm]{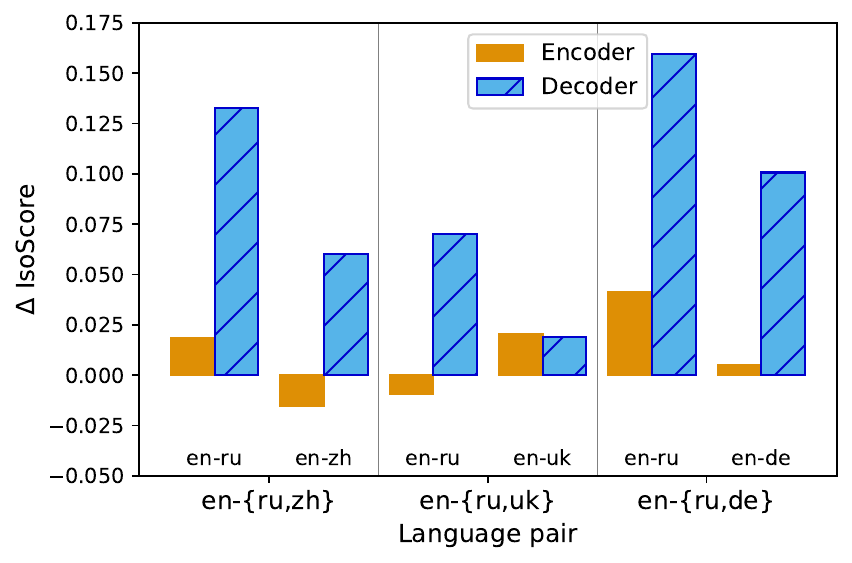} }} \\
    \subfloat[\centering Multiparallel TED]{{\includegraphics[width=7cm]{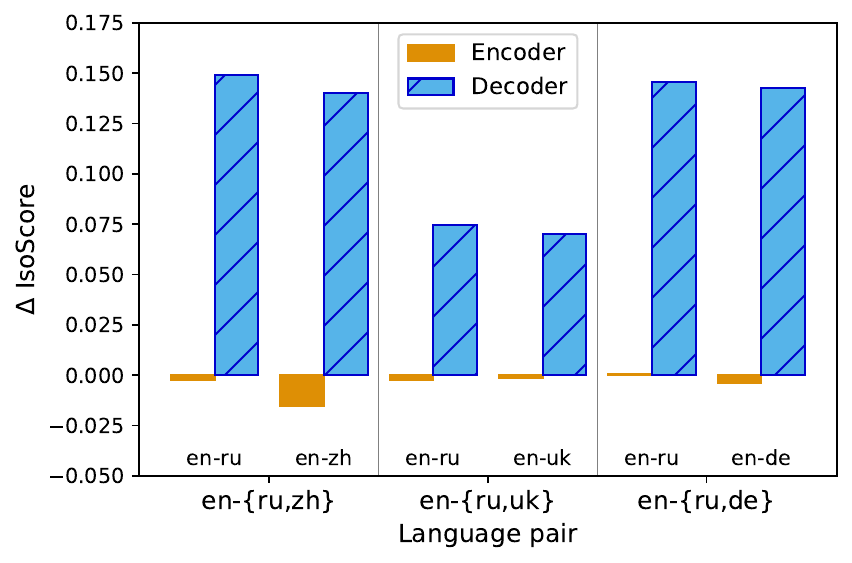} }}
     \caption{$\Delta$IsoScore values between language-specific multilingual representations separated by language and overall multilingual representations, for both the encoder and decoder ($Iso(X^{\text{multi}}(s, t_k) - Iso(X^{\text{multi}}(s, \cup_j t_j) $). Large $\Delta$IsoScores between language-specific multilingual reps. and overall multilingual reps. indicate heavy encoding of language specificity in the decoder space. 
     }
    \label{fig:mult_separation}%
\end{figure}

\subsection{Decoder language separation}
\label{sec:dec_lang_sep}
Across all three language settings, and in all of our data settings, we see that the isotropy of the overall multilingual decoder hidden space is much lower than either of the specific language portions of the multilingual space.
What this suggests, according to our metrics, is that there are some dimensions whose variance is heavily dictated by language information. 
When separating out these representations by language, the variance is reduced. This, however, is not the case when considering encoder language separation.
We summarize this phenomenon in Figure \ref{fig:mult_separation}. 

In our multiparallel setting, tested on both our TED and WMT datasets, we see that this difference is smallest for en-\{ru,uk\}. 
We hypothesize that this difference is due to vocabulary sharing. 
Because Russian and Ukrainian share a script and subword units, shared output embedding vocabulary items would lead to closer hidden states. Their close typological relatedness could be contributing to their decoders state closeness as well. 
However, since Russian and German or Russian and Chinese share very few vocabulary units, their hidden states are further in the multilingual decoder space, as also seen in Figure \ref{fig:mult_separation}.

\subsubsection{Layerwise decoder behavior}

\begin{figure*}[ht!]%
    \centering
    \subfloat[\centering en-\{ru,zh\}]{{\includegraphics[width=0.33\textwidth]{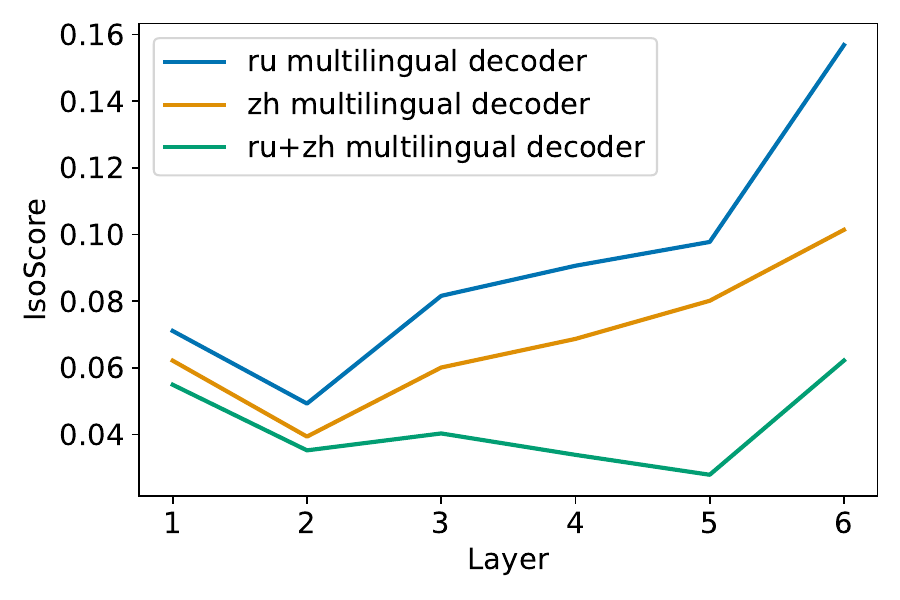} }}%
    \subfloat[\centering en-\{ru,uk\}]{{\includegraphics[width=0.33\textwidth]{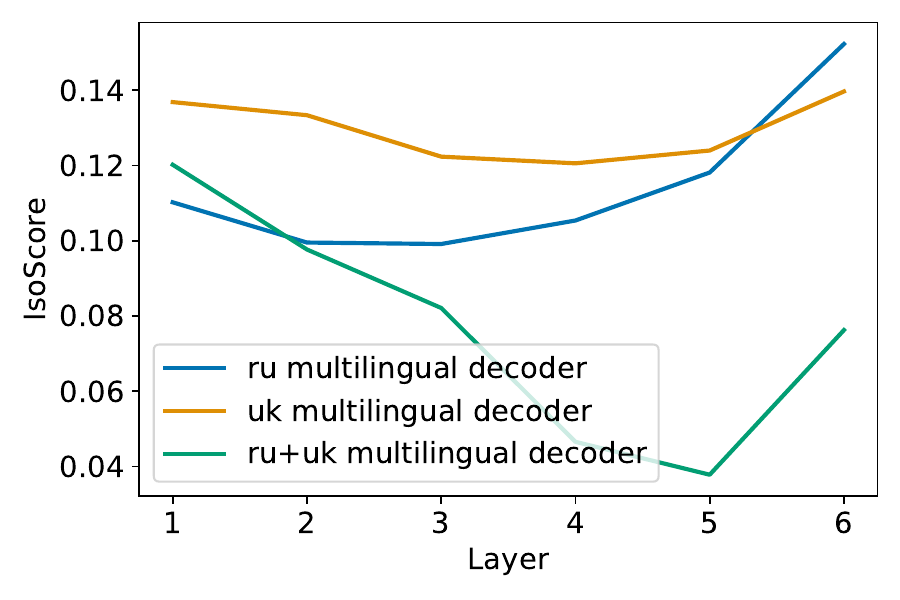} }} 
    \subfloat[\centering en-\{ru,de\}]{{\includegraphics[width=0.33\textwidth]{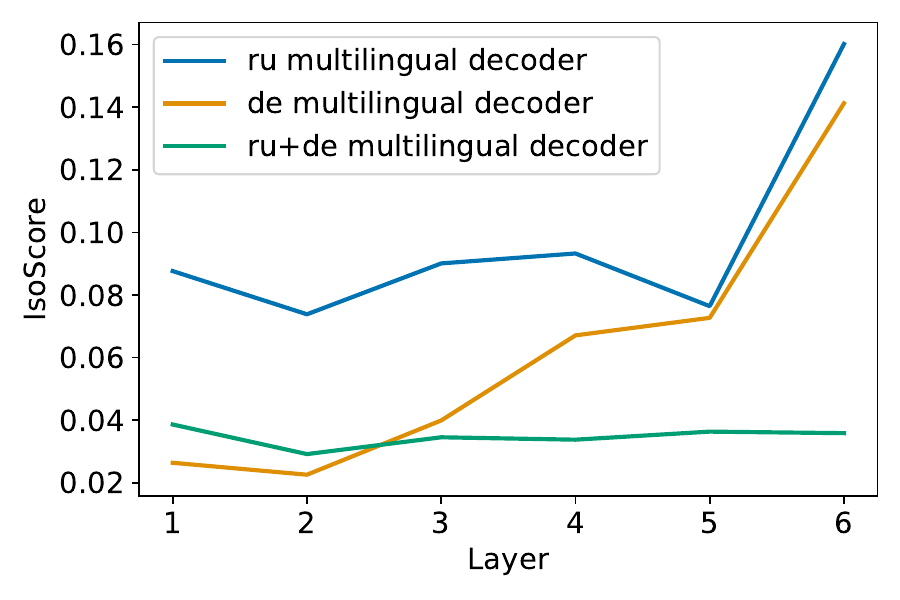} }}
    \caption{Layerwise IsoScores on our WMT-large models. 
    The divergence between the overall decoder isotropy and language-specific isotropy shows that hidden states become more language-specific throughout the decoder. }%
    \label{fig:layerwise}%
\end{figure*}

We further investigate our claim that multilingual decoders use significant representational capacity to model language-specific information by observing how isotropy changes in multilingual decoder states across decoder layers. We show layer-wise isotropy results for multilingual decoder states in Figure \ref {fig:layerwise}. We obtain hidden states according method described in Section \ref{sec:computing_isotropy}, but instead at each layer boundary. 

We find that throughout decoder layers, the overall isotropy of the entire set of decoder hidden states remains constant or decreases. However, for language-specific decoder states, we see that isotropy increases throughout the layers. Together, this implies that throughout the decoder layers, representations become more language specific. 
This suggests that earlier layers in the decoder benefit from some sharing, whereas later layers handle greater language specificity.

 In summary, these results seem to suggest that decoders in multilingual translation models seem to separate out languages among the dimensions available in their hidden states. This finding could motivate the design and use of multilingual architectures that do not use complete sharing in their decoder parameters. Some prior work has already examined this approach \cite{sachan-neubig-2018-parameter, kong-etal-2021-multilingual, nllbteam2022language}.

\section{Related Work}

\subsection{Multilingual model capacity}

Prior work has also examined the bottleneck phenomenon in multilingual machine translation. Much of this work observes the phenomenon empirically, and proposes methods to try to alleviate the parity. \citet{sachan-neubig-2018-parameter} also focus on one-to-many translation models, and propose partial sharing between language decoders in order to reduce the observed interference during full sharing. \citet{tanmultilingual} propose a knowledge distillation method to reduce the parity between bilingual and multilingual translation models by using bilingual models as multiple teachers and the multilingual model as a student.  Other methods propose using a mix of language-specific and language-agnostic parameters, \cite{lin-etal-2021-learning} and even automatically learning where to and where not to share across language pairs \cite{Zhang2021ShareON}. \citet{wanggradient} approach interference from a gradient viewpoint, and find that in En$\rightarrow$ Any models, gradients become less similar in decoders, and hypothesize that this is due to the difference in decoder label spaces. 

\citet{kudugunta-etal-2019-investigating}, like us, also investigate hidden representations to understand sharing in multilingual translation models. However, they focus on an in-house many-to-many translation model, and focus on representational similarities between languages, rather than  representational capacity for language pairs.  

\citet{shaham_causes} take an empirical approach to understanding interference in multilingual translation models, by investigate how scale and multilingual dataset ratios affect performance. 
They propose to both scale up models and adjust temperature sampling to reduce interference for simple models. 
However, this approach is largely empirical, and does not account for smaller scales and balanced datasets.

\subsection{Isotropy of Representations}

Recently, studies analyzing the geometry of Transformer representations have shown that they do not uniformly occupy many of the dimensions of the underlying space in which they lie. 
\citet{ethayarajh-2019-contextual} show that many pretrained language models are anisotropic, where any two representations have very high cosine similarity. In addition to proposing a new metric, \citet{rudman-etal-2022-isoscore} also find that in their revised analysis, representations from language models use even fewer dimensions than previously reported. 
In the translation setting, \citet{gaorepresentation} show that embeddings from generation models, including MT models, tend to degenerate into an anisotropic distribution due to frequency bias. \citet{yu-etal-2022-rare} find a similar degeneration in generation models, and propose a gradient gating method that helps reduce the frequency bias causing embedding isotropy. They report improved MT results when controlling for anisotropy.

\section{Conclusion}

While previous work has empirically demonstrated performance differences in multilingual and bilingual models, in this work, we systematically compare the geometry of model representations in bilingual and multilingual translation models in order to determine what might drive these differences.
Using one-to-many models which are most prone to interference, we experiment with varying data sizes and source-target combinations.

We find for a given language pair, there is a consistent reduction in representational capacity in multilingual decoders versus comparable bilingual decoders. 
We additionally find a small increase in representational capacity for multilingual encoder spaces given the one-to-many task. 
Representational capacity decreases in a larger model and data paradigm, and results on multiparallel data show a strong improvement in multilingual encoder representational capacity and some improvement in multilingual decoder representational capacity.  
Finally, we find that reduced capacity in multilingual decoders can be attributed to language information occupying a significant portion of the available representation space.

\section{Limitations}

Our models cover at most 3 language families for the sake of controlled analysis when modern multilingual translation models cover many more. 
We think it is worthwhile to analyze models with larger coverage as future work. 
We focus on one-to-many models as they tend to fall behind other multilingual model types \cite{sachan-neubig-2018-parameter, wang-etal-2018-three, shaham_causes}. However, many-to-many models still have multilingual decoders but may have different behavior given their multilingual encoder state space.

Additionally, our conclusions focus on encoder-decoder models, but there is growing interest in decoder-only translation models whose isotropic behavior may differ. 

Finally, our work focuses only on the characterization of representational capacity differences between model types, and not on the improvement of representational capacity of one-to-many models. However, we hope this work provides insight into the development of future modeling techniques for models with multilingual decoders.

\nocite{*}
\section{Bibliographical References}\label{sec:reference}

\bibliographystyle{lrec-coling2024-natbib}
\bibliography{lrec-coling2024-example}

\newpage
\appendix

\section{WMT Data Preprocessing}
\label{sec:wmt_cleaning_appendix}

We preprocess and filter the WMT training data in order to ensure a set of high quality bitext from the original crawled data provided by organizers. Steps 1-4 are reproduced from \citet{m2m100}. 

\begin{enumerate}
\item Remove lines that are > 50\% punctuation
\item Deduplicate training data
\item Language-specific filtering to remove sentences that are > 50\% characters that are not identified as belonging to the given language. 
\item Length ratio cleaning with ratio=3, and remove sentences with > 250 subwords. 
\item Language identification filter such that both the source and target language ID must be correct. We use the \texttt{fasttext} LangID model: $\texttt{lid.176.bin}$. \cite{joulin2016fasttext, joulin-etal-2017-bag}. 
\item Bitext filtering using LASER Embeddings as implemented by the $\texttt{OpusFilter}$ toolkit \cite{aulamo-etal-2020-opusfilter, artetxe-schwenk-2019-margin}. 
\end{enumerate}

\newpage
\section{TED Models on WMT dev set}
\label{sec:appendix_ted_wmt}

\begin{table}[ht!]
\begin{center}
\begin{tabular}{lcccc}
\toprule
& &\multicolumn{3}{c}{TED} \\
\cmidrule(lr){3-5}
 langs & type & BLEU & iso-enc  & iso-dec \\
\midrule  
\multirow{2}{*}{en-ru} & multi & 10.6 & 0.102 & 0.227  \\
& bi &  10.3 &\textbf{0.107} & \textbf{0.264} \\
\multirow{2}{*}{en-zh}  & multi & 16.4  & \textbf{0.084} & 0.166 \\
&bi & 15.3 & 0.034 & \textbf{0.194} \\
 & multi & - & 0.092 & 0.056 \\
\midrule
 \multirow{2}{*}{en-ru} & multi & 11.0  & \textbf{0.097 }& \textbf{0.243} \\
& bi & 10.2 & 0.076 & 0.235\\
 \multirow{2}{*}{en-de}  & multi & 16.5 & \textbf{0.073} & 0.223 \\
& bi & 15.2 & 0.040 &\textbf{ 0.227 }\\
& multi & -& 0.079 & 0.085  \\
\midrule
\multirow{2}{*}{en-ru} & multi  & 7.7 & \textbf{0.125} & \textbf{0.228}\\
& bi & 7.1 & 0.103 & 0.213\\
\multirow{2}{*}{en-uk} & multi &  11.2 & \textbf{ 0.143} &\textbf{ 0.202}\\
& bi & 10.0 & 0.131 & 0.188\\
 & multi & - & 0.130 & 0.174 \\
\bottomrule
\end{tabular}
\end{center}
\label{table:ted_crosstest_wmt}
\caption{Isotropy results on our multiparallel TED model, tested on the WMT development set for direct comparison with our other models.}
\end{table}

\begin{figure}[h]
    \centering
    \includegraphics[width=\columnwidth]{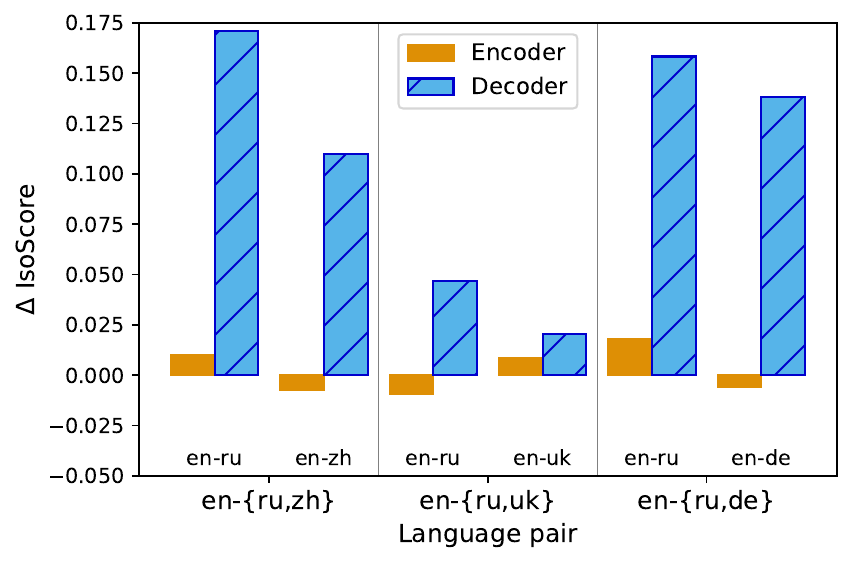}
    \caption{$\Delta$ IsoScore values between language-specific multilingual representations separated by language and overall multilingual representations, for both the encoder and decoder.}
    \label{fig:page-four}
\end{figure}

\end{document}